\documentclass[10pt, a4paper]{article}

\usepackage{lrec2026}

\usepackage{graphicx}
\usepackage{amsmath,amssymb,amsfonts}
\usepackage{algorithmic}
\usepackage{textcomp}
\usepackage{booktabs}
\usepackage{multirow}
\usepackage[nolist]{acronym}
\usepackage{enumitem}
\usepackage{cmap}
\usepackage{url}
\usepackage{siunitx}
\usepackage{caption}
\usepackage{subcaption}
\usepackage{balance}

\setlist{itemsep=1pt, topsep=2pt, parsep=0pt, partopsep=0pt}

\title{Quantifying the Accuracy and Cost Impact of Design Decisions in Budget-Constrained Agentic LLM Search}

\name{Kyle A. McCleary and James M. Ghawaly}
\address{Division of Computer Science \& Engineering, Louisiana State University\\
Baton Rouge, Louisiana, USA\\
jghawaly@lsu.edu, kmccl24@lsu.edu}

\abstract{Agentic Retrieval-Augmented Generation (RAG) systems combine iterative search, planning prompts, and retrieval backends, but deployed settings impose explicit budgets on tool calls and completion tokens. We present a controlled measurement study of how search depth, retrieval strategy, and completion budget affect accuracy and cost under fixed constraints. Using \emph{Budget-Constrained Agentic Search (BCAS)}, a model-agnostic evaluation harness that surfaces remaining budget and gates tool use, we run comparisons across six LLMs and three question-answering benchmarks. Across models and datasets, accuracy improves with additional searches up to a small cap, hybrid lexical and dense retrieval with lightweight re-ranking produces the largest average gains in our ablation grid, and larger completion budgets are most helpful on HotpotQA-style synthesis. These results provide practical guidance for configuring budgeted agentic retrieval pipelines and are accompanied by reproducible prompts and evaluation settings.\\ \newline \Keywords{Retrieval-Augmented Generation, Agentic Search, Budget-Aware Evaluation}}

\begin{document}

\maketitleabstract

\begin{acronym}
\acro{BCAS}{Budget-Constrained Agentic Search}
\acro{IR}{Information Retrieval}
\acro{LLM}{Large Language Model}
\acro{RAG}{Retrieval-Augmented Generation}
\acro{RL}{Reinforcement Learning}
\acro{QA}{Question Answering}
\end{acronym}

\section{Introduction}
\acused{RAG}
The landscape of Retrieval-Augmented Generation (RAG) has undergone a significant shift in just one year. What began as static retrieve-and-generate pipelines has evolved into sophisticated agentic systems where \acp{LLM} autonomously plan and execute multi-step retrieval strategies~\cite{singh2024agentic}. Agentic retrieval capabilities are now integrated into mainstream AI tools. OpenAI's ChatGPT employs function calling for iterative search~\cite{openai2023function}, Anthropic's Claude features a dedicated research mode~\cite{anthropic2024research}, and Google's Gemini 2.0 includes native tool-use APIs for dynamic information gathering~\cite{gemini25}.

However, while these advances have primarily focused on maximizing retrieval quality and answer accuracy, a critical dimension remains underexplored: \emph{computational budget constraints}. In real-world deployments, the cost of multiple API calls, the latency of sequential searches, and the computational overhead of processing numerous retrieved documents present significant challenges. 

In this setting, we frame the problem as a measurement gap. \ac{RAG} is now the default strategy for grounding \acp{LLM}, and many systems combine iterative querying, intermediate reasoning, hybrid lexical $+$ dense search, and re-ranking under ``agentic search.'' Existing studies typically report accuracy improvements for individual methods, often with permissive tool usage, but fewer quantify how these design knobs jointly affect both quality and cost under fixed search and token budgets across multiple models and datasets.

In this work, we quantify how common agentic design decisions, including hybrid search, lightweight re-ranking, pre-planning, reflection, and completion limits, shift accuracy and cost when searches and tokens are capped. We conduct controlled comparisons across six LLMs and three multi-hop QA benchmarks, holding budgets constant while toggling one decision at a time.

To run these experiments we use \textbf{\ac{BCAS}}, a budget-aware evaluation harness that surfaces remaining search and token allowances as explicit signals and gates tool calls accordingly. BCAS keeps the instrumentation simple: it reuses commodity prompts, avoids bespoke APIs, and records per-question search and token consumption so practitioners can reinterpret the results under their own pricing models.

\subsection{Research Questions}

To quantify the practical value of budget-aware planning, we subject \ac{BCAS} to a large-scale study involving six LLMs and three multi-hop benchmarks, guided by three questions:

\begin{enumerate}[
        label=\textbf{RQ\arabic*:},
        leftmargin=*,
        labelsep=0.6em,
        itemsep=0.2em,
        align=left]   
    \item \textbf{Performance Across Model Sizes}\\
            \textit{How does model size (parameter count) impact retrieval performance?}
    \item \textbf{Budget-Aware Component Tuning}\\
          \textit{How do individual agentic-RAG components and hyper-parameters influence performance under a fixed search budget, and can those effects be exploited to optimize performance within that budget?}
    \item \textbf{Accuracy-Budget Trade-off}\\
          \textit{How does accuracy vary as we simultaneously tighten the budgets on iterative search steps \emph{and} generated output tokens in an agentic-RAG pipeline?}
\end{enumerate}

\noindent Recent reinforcement-learning-driven systems such as Search-R1~\cite{jin2025searchr1} and DeepRetrieval~\cite{jiang2025deepretrieval} demonstrate large gains but incur significant training and inference overheads, so cost-aware behaviour in agentic RAG warrants closer study.
\noindent The remainder of the paper reviews related work in Section~2, documents the evaluation protocol and budget controls in Section~3, presents results in Section~4, and concludes with discussion and takeaways in Sections~5 and~6.

\section{Related Work}\label{sec:related_works}

Agentic retrieval has progressed from static pipelines to autonomous research agents within a few years, providing the backdrop for \ac{BCAS}.

\noindent\textbf{Static pipelines.}\space Early systems such as RAG~\cite{lewis2020rag} and FiD~\cite{izacard2021fid} established the retrieve-then-generate paradigm and demonstrated how parametric and non-parametric memories could be combined effectively.

\noindent\textbf{Prompted agency.}\space Lightweight prompting soon enabled iterative reasoning without new training. ReAct~\cite{yao2023react} introduced the ``Thought, Action, Observation'' loop for search planning, Self-Ask~\cite{press2023selfask} decomposed questions into sub-queries, and FLARE~\cite{jiang2023flare} triggered retrieval on demand. Reflection-driven methods such as Self-RAG~\cite{asai2024selfrag} and CRAG~\cite{yan2024crag} further improved robustness by critiquing intermediate results.

\noindent\textbf{Autonomous research agents.}\space Recent work deepens external planning through structured reasoning and reinforcement learning. Self-Reasoning~\cite{xia2025selfreasoning} trains models to analyse their own trajectories, while Search-R1~\cite{jin2025searchr1} and DeepRetrieval~\cite{jiang2025deepretrieval} learn to formulate and schedule queries directly against search engines.

\noindent\textbf{Budget-aware gap.}\space These approaches often assume generous tool usage and many depend on bespoke training or fine-tuning. \ac{BCAS} contributes an orthogonal evaluation layer: it makes compute constraints explicit at inference time, standardizes the retrieval interface, and logs cost telemetry so design choices can be compared under shared budgets. We quantify how search depth, component choices, and token limits interact across diverse \acp{LLM} in this controlled setting.

This distinction matters for how the paper should be read. We do not introduce a new trained search agent or claim a new state of the art on any benchmark. Instead, we use a common scaffold to isolate which design choices matter once searches and completion length are capped. That makes the contribution closer to controlled systems measurement than to agent leaderboard optimization.


\begin{figure*}[t]
\centering
\includegraphics[width=\textwidth,
                 trim=2.8cm 0 0 0,clip]{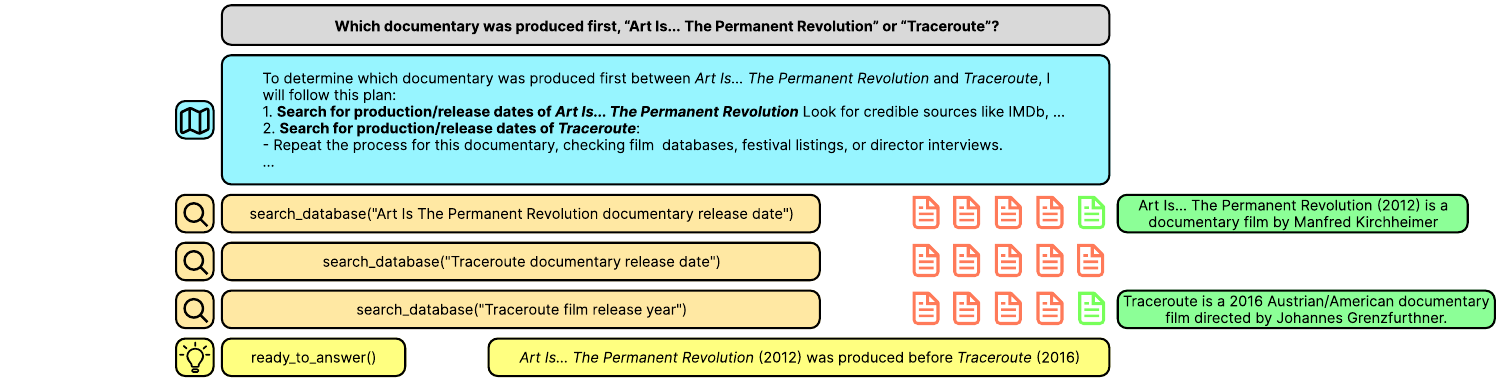}
\caption{An example of DeepSeek V3 (0324) solving a HotpotQA `hard' problem using \ac{BCAS} with 4 searches and planning. This demonstrates the core principles of BCAS: the model operates within a strict 4-search budget (RQ1, RQ3) while leveraging system components like planning and iterative search to efficiently locate the answer (RQ2) and terminate early.}
\label{fig:trajectory}
\end{figure*}

\section{Methodology}\label{sec:methodology}

This section provides implementation details for the \ac{BCAS} framework, describes the BCAS component ablation studies, and also the iterative search count and output context size scaling studies. We use this loop only as a controlled evaluation scaffold to quantify how search depth, retrieval components, and token limits affect accuracy and cost under explicit budgets. An example of \ac{BCAS} in action is shown in Figure~\ref{fig:trajectory}, where DeepSeek V3 (0324) solves a complex HotpotQA question within a strict search budget. Our implementation and evaluation code is available at 
\url{https://github.com/kmccleary3301/BCAS_RAG}.


\subsection{Architectural Principles}
BCAS is built on three design principles that align with our research questions:

\noindent\textbf{Explicit budget management (RQ3).} Models receive their remaining search and token allowances at each step, nudging them toward deliberate query sequencing. The primary levers are \verb|max_searches| and \verb|max_total_tokens|.

\noindent\textbf{Composable component pipeline (RQ2).} Retrieval tools, planning hooks, and reflection routines are toggled through configuration, enabling controlled ablations of each capability.

\noindent\textbf{Model-agnostic design (RQ1).} Prompts describe tools in plain language rather than bespoke APIs, so any instruction-following \ac{LLM} can adopt the framework without fine-tuning.

\subsection{Prompt Policy}
For each evaluated method, we use one prompt template shared across all model families, and we do not tune prompts per model. This choice keeps the interaction protocol fixed so component and budget effects can be compared directly. As a result, cross-model differences in this study should be interpreted as sensitivity to a shared scaffold rather than each model's best-case tuned performance.

\subsection{Budget Accounting}
BCAS tracks two resources throughout each trajectory: remaining search calls and cumulative completion budget. The search cap is enforced as a hard tool constraint. Once the allowance is exhausted, the search tool is removed from the available action list and the model must answer with the information it has already gathered. Token accounting is updated after every turn from the provider-reported usage fields, and the loop stops once the configured \verb|max_total_tokens| threshold is reached. This accounting scheme is deliberately simple. It does not model queueing delays, batching effects, or provider-specific pricing tiers, but it does give a consistent view of how each configuration converts a fixed interaction budget into answer accuracy.

\subsection{The BCAS Execution Loop}
The BCAS algorithm is implemented as a stateful loop that guides the LLM through a budget-aware reasoning and retrieval process. In each turn, the model:
\begin{enumerate}[
    label=\textbf{\arabic*:}
]
    \item \textbf{Reasons:} Generates a ``thought" about its current progress and what information it needs next.
    \item \textbf{Selects a Tool:} Chooses an action from a dynamically generated list of available tools. The \verb|search_database| tool is only available if the agent is still within its \verb|max_searches| budget.
    \item \textbf{Executes and Observes:} The chosen tool is executed (e.g., a database search is performed), and the result (an ``observation") is returned to the model.
    \item \textbf{Updates Budget:} Token usage is updated, and the remaining budget is recalculated for the next turn.
\end{enumerate}
This loop continues until the model calls the \verb|ready_to_answer| tool or a budget is exhausted.

\paragraph{Optional Pre-planning} When enabled, the agent decomposes the question into a step-by-step research plan, considering available search and token budgets for more structured approaches to multi-hop questions.

\paragraph{Optional Reflection} This periodically prompts the model to review progress and make strategic adjustments, particularly effective for smaller models in recognizing unproductive search paths and optimizing budget usage.

These planning and reflection strategies are implemented following patterns from recent state-of-the-art works on agentic search~\cite{xia2025selfreasoning,yao2023react,asai2024selfrag}, adapted to operate within explicit budget constraints.

\subsection{Information Retrieval Components}
To analyze component tuning (RQ2), we equip the agent with three distinct \ac{IR} strategies that can be configured for each experiment.

\begin{itemize}
    \item \textbf{BM25 with Phrase Boosting:} Our baseline retriever uses a standard BM25 algorithm enhanced with a query parser that boosts two- and three-term phrases. This prioritizes documents with dense coverage of the query's concepts, inspired by cover density~\cite{Cover_Density}.
    \item \textbf{Hybrid Search (HS):} This method combines the BM25 retriever with dense vector search,
    offering a blend of lexical and semantic matching.
    \item \textbf{Re-ranking (RR):} We use a context re-ranking model to re-score and select the top five most relevant results. This can be applied on top of either BM25 or Hybrid Search. 
\end{itemize}

Our \ac{IR} backend is built on ParadeDB~\cite{paradedbgithub}, a PostgreSQL extension. This allows us to construct a hybrid BM25 and vector search engine over our documents.

\paragraph{Retrieval and re-ranking pipeline.}
Each search request returns chunk-level candidates from the dataset-specific index. In BM25 mode, the system returns the top five BM25 chunks. In hybrid mode (\textbf{HS}), lexical and dense retrieval are combined and the top five chunks are returned to the agent context. In hybrid + re-ranking mode (\textbf{HS+RR}), we first retrieve 100 hybrid candidates, then re-score them with a cross-encoder re-ranker and keep the top five chunks for the model context.

We keep the final context window deliberately narrow at five returned chunks so that search policy, rather than indiscriminate context expansion, remains the main source of improvement. The 100-candidate pool in the re-ranking condition is large enough to expose meaningful ordering differences while remaining cheap enough to run across the full ablation grid.

\subsubsection{Vector Embedding}
We used the \verb|BGE-M3|~\cite{bge-m3} model for embedding document chunks and search queries during hybrid retrieval. \verb|BGE-M3| was chosen for its broad effectiveness in multilingual benchmarks, as outlined in~\cite{bge-m3}. Vector similarity was measured using the standard cosine similarity metric. In our results, we denote hybrid retriever search as \textbf{HS}.

\subsubsection{Re-ranking}
We tested the effects of re-ranking results at retrieval. We used \verb|bge-reranker-v2-m3| for this task due to competitive performance on common benchmarks~\cite{baai_rerank}. During testing cases involving the re-ranker, we retrieved 100 document chunks using hybrid retrieval, calculated scores for these chunks using the re-rank model, and returned the top five chunks. In our results, we denote re-ranking as \textbf{RR}. 

\subsection{Evaluation}
We evaluate \ac{BCAS} across six \acp{LLM} and three benchmarks to measure how accuracy scales with search depth and completion token count. We also evaluate the effects of various retrieval components on the accuracy of unconstrained \ac{BCAS} search through ablation.

\subsubsection{LLM Selection}
We evaluated six models across different size categories to analyze performance across model capacities (RQ1): \textbf{o4-mini}~\cite{o3_System_Card}, \textbf{DeepSeek V3 (0324 version)}~\cite{deepseek2024}, \textbf{GPT-4.1-mini}~\cite{GPT4_1_Announcement}, \textbf{Gemma 3 27B}~\cite{gemma2}, \textbf{Qwen 3 14B}~\cite{qwen2024}, and \textbf{LLaMA 3.1 8B}~\cite{Llama3_2024}. These models represent diverse capabilities, costs, and architectures, all proficient in tool usage for BCAS tasks.

\subsubsection{Grading}
We use GPT-4o-mini as a binary correctness judge, following prior evidence that LLM judges can track human answer grading reasonably well in QA settings~\cite{judgingllmasajudgemtbenchchatbot}. The judge receives the question, reference answer, and model answer, and outputs a correct/incorrect label under a constrained rubric. GPT-4o-mini does not, to our knowledge, share a model family with the six evaluated answering models.
In addition to literature support, we conducted an independent manual audit of 200 TriviaQA samples, 200 ``hard" HotpotQA samples, and 200 2WikiMultihopQA samples. We confirmed correct LLM grading on 200/200 TriviaQA, 198/200 HotpotQA, and 197/200 2WikiMultihopQA samples. The five discrepancies were all false negatives from the judge, so reported accuracies are slightly conservative. We still treat binary judge labels as a proxy metric and discuss this limitation in Section~7.

\subsection{Datasets}\label{sec:datasets}

We performed our scaling evaluations on three diverse question-answering benchmarks: 

\begin{enumerate}[
        label=\textbf{\arabic*:},
        leftmargin=*,
        labelsep=0.6em,
        itemsep=0.2em,
        align=left]   
    \item \textbf{TriviaQA}~\cite{TriviaQA_2017} is a widely used benchmark for open-domain question-answering that requires finding specific facts, often from a single document. We use the unfiltered test set, comprising over 500K documents.
    \item \textbf{HotpotQA}~\cite{HotpotQA_2018} focuses on multi-hop reasoning, where questions require synthesizing information from multiple documents to arrive at an answer.
    \item \textbf{2WikiMultihopQA}~\cite{2wikimultihop} is the most difficult of the three, demanding reasoning over multiple hops with more complex entity relationships.
\end{enumerate}

These datasets were chosen to represent a range of complexities, from single-fact retrieval to complex multi-step reasoning. Each sample contains a question, answer, and relevant documents.

The documents for each dataset were partitioned into separate collections, and retrieval was restricted to the appropriate collection during testing.
For all cases, we ensured that each model was evaluated on the same set of samples, verified by hashing the dataset before evaluation. Due to occasional API failures during evaluation, final sample counts ranged from 467 to 537 samples across datasets, with counts held constant within each dataset.
Each run logs per-question metadata, including \verb|searches_used|, \verb|tokens_in|, \verb|tokens_out|, \verb|early_stop|, and grader verdicts, along with a cost ledger so practitioners can recompute dollar metrics under alternative price sheets.

\subsection{Ablation Study}
We perform an ablation study of \ac{BCAS} to measure the effect of optional features and altered search tooling on retrieval accuracy. This was performed on all six models using 467 testing samples from HotpotQA (samples released with our code). Our baseline configuration has \verb|max_total_tokens| set to 16000, \verb|max_searches| set to ``unlimited'', the search endpoint set to BM25 only, pre-planning and reflection disabled, and context + search hints enabled. All of our feature adjustments are shown by their net effect on the baseline configuration's accuracy. We measure the effect of the following:
\begin{enumerate}
    \item Pre-planning
    \item Reflection (no Pre-planning)
    \item Pre-planning \textit{and} Reflection
    \item Hybrid BM25 + Vector Search (n=5).
    \item Hybrid BM25 + Vector Search (n=100) with re-ranking of results (choose top 5).
\end{enumerate}

As discussed in the limitations, we report the full factorial ablation only on HotpotQA, where multi-hop synthesis makes component interactions most visible. We do not claim that the exact ordering of ablation gains is universal across datasets.

\subsection{Search Scaling Study}
To understand how iterative search capacity affects performance (RQ3), we systematically varied the \verb|max_searches| parameter across four configurations: 1, 2, 3, and unlimited searches. All experiments used hybrid search returning 5 results per query with \verb|max_total_tokens| = 16,000, and both preplanning and reflection disabled. This standardized baseline ensures that observed performance differences are attributable to search budget variations rather than confounding factors. Additionally, we evaluated two enhanced configurations using unlimited search: with preplanning enabled, and with both preplanning and reflection enabled. This design yields six total configurations per model-dataset combination, allowing us to trace both the core search scaling effect and its interaction with strategic reasoning components.

\subsection{Context Scaling Study}
To examine how token budget constraints influence agent behavior (RQ3), we evaluated performance across five context limits: 500, 1K, 2K, 4K, and 16K tokens using hybrid search with unlimited searches, ensuring that performance variations reflect context budget effects rather than search limitations. We excluded o4-mini due to its reasoning model architecture, where completion tokens include the internal reasoning trace, making artificially low token limits likely to truncate responses mid-reasoning rather than demonstrating budget-aware behavior. Enhanced reasoning configurations are included as appropriate and correspond to identical conditions from the search scaling study. Thus comparisons across context limits exclude o4-mini to avoid conflating budget policy with model-internal reasoning tokenization.

\section{Results}\label{sec:results}

\begin{figure*}[t]
\centering
\includegraphics[width=\textwidth]{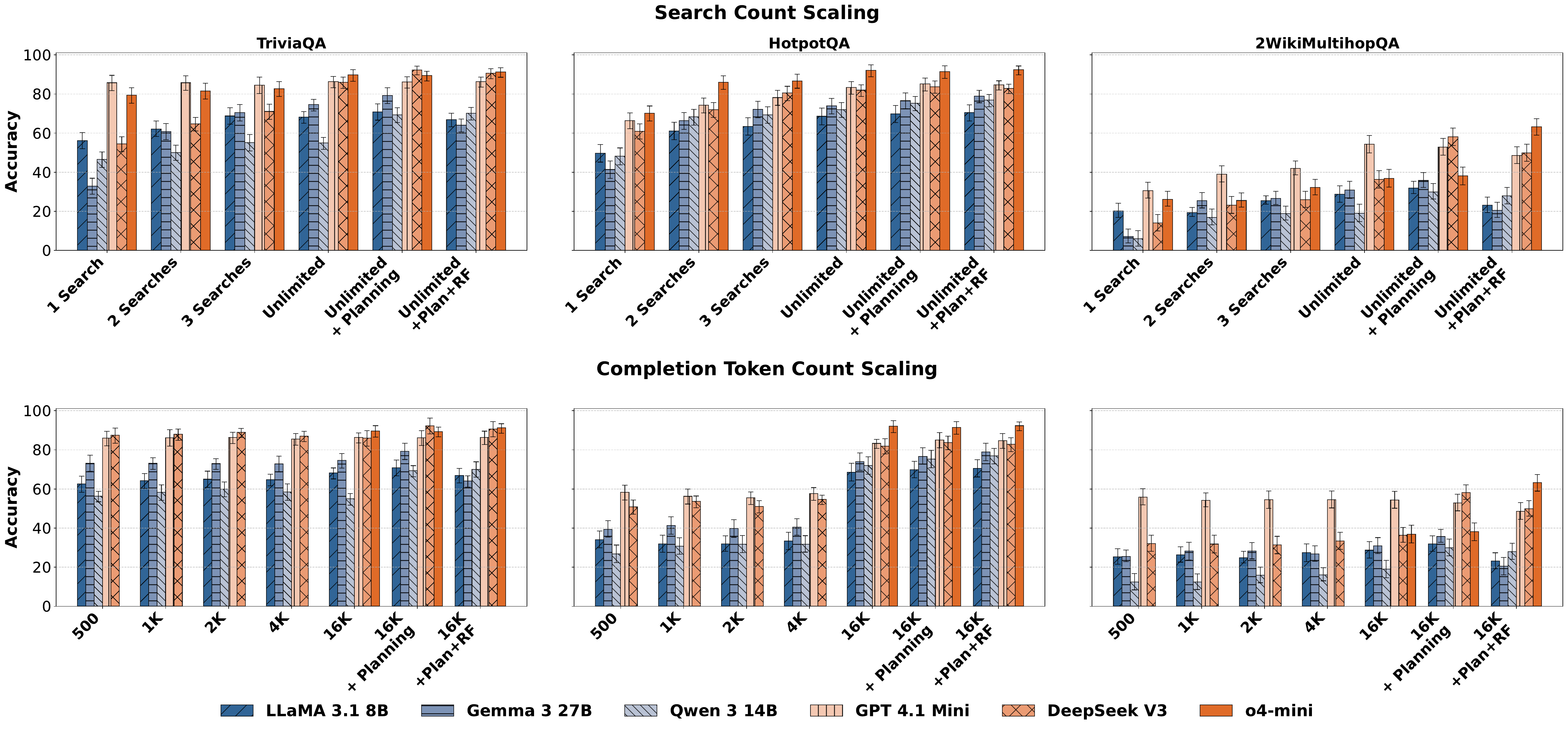}
\caption{Search scaling and context scaling performance across TriviaQA, HotpotQA, and 2WikiMultihopQA. The "Planning" columns show the impact of adding pre-planning and reflection (RF) strategies. This figure provides the primary data for analyzing the accuracy-budget trade-off (RQ3). Confidence intervals computed using 95\% Wilson score interval~\cite{wilson1927probable}.}
\label{fig:all_scaling_data}
\end{figure*}

\begin{figure}[t]
\centering
\includegraphics[width=\columnwidth]{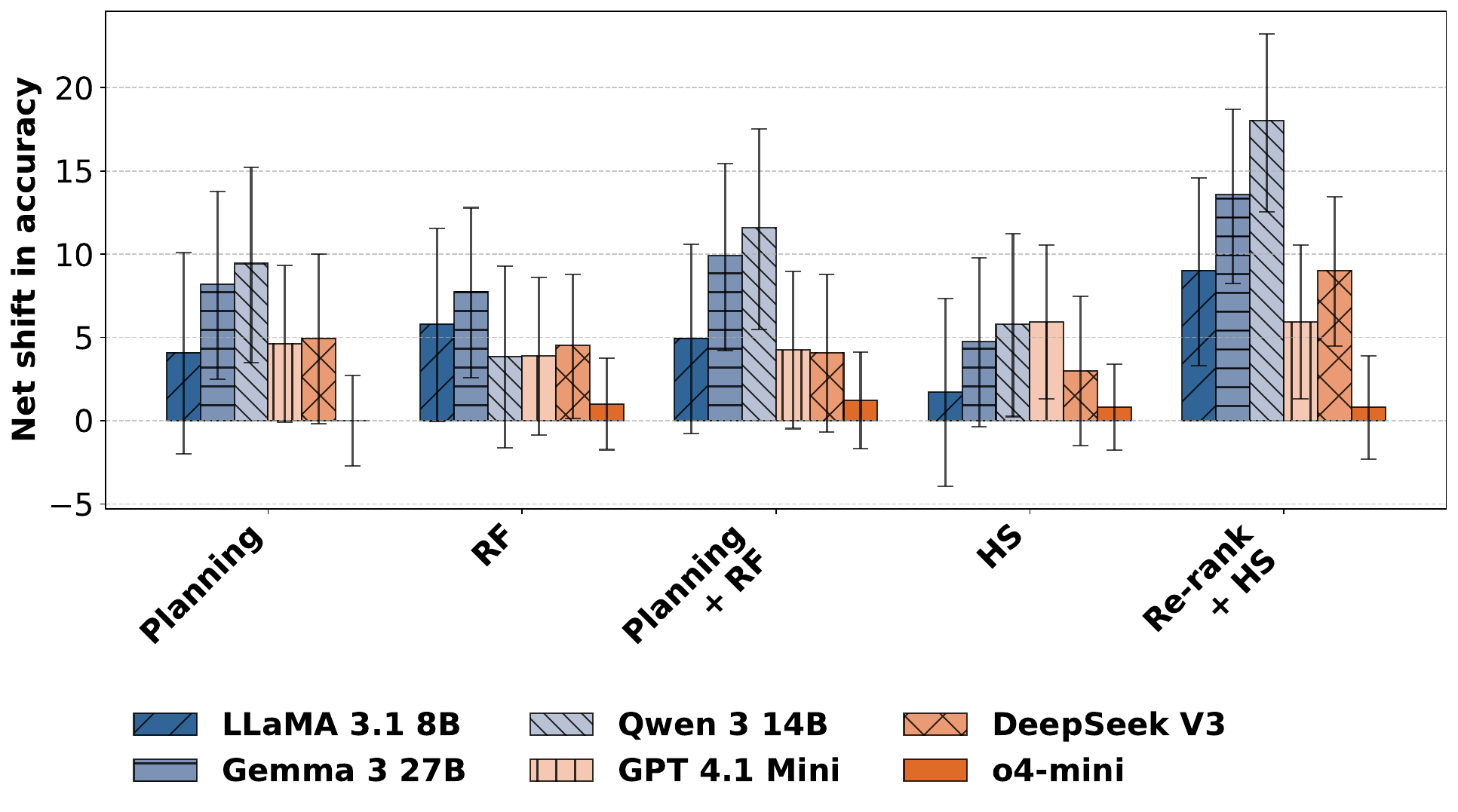}
\caption{Ablation study of BCAS features on HotpotQA: We measure the net effect on retrieval accuracy of different features compared to BCAS baseline with a BM25 retriever. 95\% confidence intervals computed using Newcombe Method 9~\cite{newcombe1998paired}.}
\label{fig:ablation}
\end{figure}

We find three consistent patterns across models and datasets: (i) accuracy improves reliably up to about three searches; (ii) hybrid retrieval with re-ranking yields the most consistent gains among components; and (iii) larger token generation budgets primarily help multi-hop questions with challenging information synthesis requirements. Figure~\ref{fig:all_scaling_data} displays the scaling trends; Figure~\ref{fig:ablation} displays the ablation study results; Table~\ref{tab:scaling_metrics} reports scaling metrics and Table~\ref{tab:ablation_metrics} details the component ablations, addressing RQ1 through RQ3.

\begin{table*}[t]
\centering
\scriptsize
\setlength{\tabcolsep}{6.5pt}
\renewcommand{\arraystretch}{0.98}
\begin{tabular}{|l|rrrr|rrrr|r|r|} 
\hline
\multirow{2}{*}{\textbf{Model}} & \multicolumn{4}{c|}{\textbf{Completion Tokens}} & \multicolumn{4}{c|}{\textbf{Searches (16K Token Limit)}} & \multirow{2}{*}{\parbox{1.1cm}{\centering\textbf{Plan}}} & \multirow{2}{*}{\parbox{1.6cm}{\centering\textbf{Plan+RF}}} \\
\cline{2-9}
& \textbf{500} & \textbf{1K} & \textbf{2K} & \textbf{4K} & \textbf{1} & \textbf{2} & \textbf{3} & \textbf{Unlimited} & & \\
\hline
\multicolumn{11}{|c|}{\textbf{TriviaQA}} \\
\hline
o4-mini & - & - & - & - & 79.33 & 81.56 & 82.68 & 89.72 & 89.39 & \textbf{91.25} \\
DeepSeek V3 & 87.45 & 88.01 & 88.95 & 87.08 & 54.51 & 64.78 & 71.07 & 85.98 & \textbf{92.34} & 90.65 \\
GPT-4.1-mini & 85.96 & 86.14 & 86.33 & 85.58 & 85.74 & 85.74 & 84.49 & \textbf{86.36} & 86.17 & 86.36 \\
Gemma 3 27B & 73.22 & 73.22 & 73.03 & 72.85 & 32.91 & 60.80 & 70.44 & 74.58 & \textbf{79.25} & 64.11 \\
Qwen 3 14B & 56.37 & 58.24 & 59.93 & 58.43 & 46.54 & 50.10 & 55.14 & 54.95 & 69.35 & \textbf{70.09} \\
LLaMA 3.1 8B & 62.55 & 64.23 & 64.98 & 64.79 & 56.18 & 62.05 & 68.76 & 68.22 & \textbf{70.84} & 66.92 \\
\hline
\multicolumn{11}{|c|}{\textbf{HotpotQA}} \\
\hline
o4-mini & - & - & - & - & 70.17 & 86.05 & 86.70 & \textbf{92.92} & 91.43 & 92.36 \\
DeepSeek V3 & 50.86 & 53.65 & 51.07 & 54.72 & 60.95 & 72.02 & 80.48 & 77.94 & \textbf{83.74} & 82.87 \\
GPT-4.1-mini & 58.33 & 56.25 & 55.56 & 57.64 & 66.38 & 74.30 & 78.16 & 80.47 & \textbf{85.10} & 84.73 \\
Gemma 3 27B & 39.40 & 41.33 & 39.83 & 40.47 & 41.38 & 66.38 & 72.20 & 68.52 & 76.66 & \textbf{78.96} \\
Qwen 3 14B & 26.82 & 30.69 & 31.76 & 31.76 & 48.18 & 68.31 & 69.38 & 64.45 & 75.33 & \textbf{76.91} \\
LLaMA 3.1 8B & 34.05 & 31.91 & 31.91 & 33.40 & 49.68 & 61.03 & 63.38 & 65.31 & 69.83 & \textbf{70.52} \\
\hline
\multicolumn{11}{|c|}{\textbf{2WikiMultihopQA}} \\
\hline
o4-mini & - & - & - & - & 26.18 & 25.54 & 32.19 & 36.91 & 38.12 & \textbf{63.20} \\
DeepSeek V3 & 27.68 & 26.18 & 26.39 & 29.18 & 14.07 & 23.16 & 25.97 & 36.36 & \textbf{58.15} & 49.89 \\
GPT-4.1-mini & \textbf{55.84} & 54.11 & 54.55 & 54.55 & 30.62 & 38.97 & 41.97 & 54.39 & 52.89 & 48.61 \\
Gemma 3 27B & 25.48 & 28.27 & 28.27 & 26.77 & 7.13 & 25.49 & 26.57 & 30.89 & \textbf{35.76} & 20.56 \\
Qwen 3 14B & 12.42 & 12.42 & 11.56 & 13.28 & 6.00 & 16.92 & 18.84 & 19.06 & \textbf{29.98} & 28.66 \\
LLaMA 3.1 8B & 25.27 & 26.34 & 24.84 & 27.41 & 20.22 & 19.35 & 25.43 & 28.70 & \textbf{31.91} & 23.13 \\
\hline
\end{tabular}
\renewcommand{\arraystretch}{1.0}
\setlength{\tabcolsep}{6pt}
\caption{Full accuracy data (\%) for all models across all datasets and BCAS budget configurations. Highest values for each row are bolded. Note that for TriviaQA, the performance was nearly flat across the board. }
\label{tab:scaling_metrics}
\end{table*}

\subsection{Performance Across Model Capacities (RQ1)}
Iterative search narrows capacity gaps: smaller models approach or exceed larger models' single-search scores when allowed additional searches. For example, on HotpotQA, Qwen~3~14B with unlimited searches and planning attains 75.33\%, exceeding o4-mini's single-search 70.17\% but not surpassing o4-mini at two searches (86.05\%; Table~\ref{tab:scaling_metrics}). Similar patterns appear on TriviaQA and 2WikiMultihopQA where multi-step search raises smaller models' accuracy relative to single-step baselines (Figure~\ref{fig:all_scaling_data}).

\subsection{Budget-Aware Component Tuning (RQ2)}

On HotpotQA, re-ranking on top of hybrid search yields the largest average improvement across models (+9.29 points), with hybrid search alone also beneficial (+6.36), while planning and reflection improve smaller models by 4 to 12 points and have limited effect on o4-mini (Table~\ref{tab:ablation_metrics}). We treat these effect sizes as task-specific to HotpotQA and do not claim that the exact ordering transfers unchanged to TriviaQA or 2WikiMultihopQA.

\begin{table}[t]
\centering
\scriptsize
\setlength{\tabcolsep}{5.5pt}
\renewcommand{\arraystretch}{0.98}
\begin{tabular}{@{}lrrrrr@{}}
\toprule
\textbf{Model} & Plan & RF & Plan+RF & HS & HS+RR \\
\midrule
LLaMA 3.1 8B & +4.52 & +6.33 & +5.21 & +5.73 & +8.80 \\
Gemma 3 27B & +8.14 & +7.37 & +10.43 & +8.93 & +13.01 \\
Qwen 3 14B & +10.87 & +5.19 & +12.45 & +12.36 & +17.86 \\
GPT-4.1-mini & +4.63 & +3.89 & +4.26 & +4.07 & +5.93 \\
DeepSeek V3 & +5.79 & +4.55 & +4.92 & +6.23 & +9.02 \\
o4-mini & +0.17 & +0.92 & +1.10 & +0.83 & +1.10 \\
\midrule
Average & +5.69 & +4.71 & +6.40 & +6.36 & +9.29 \\
\bottomrule
\end{tabular}
\renewcommand{\arraystretch}{1.0}
\setlength{\tabcolsep}{6pt}
\caption{Impact of retrieval method components on HotpotQA accuracy (\%). This analysis informs component tuning under budget constraints (RQ2).}
\label{tab:ablation_metrics}
\end{table}

\subsection{Accuracy-Budget Trade-off (RQ3)}

Increasing search steps yields consistent gains up to roughly three searches, after which returns plateau; context scaling shows limited change for TriviaQA, a clear lift for HotpotQA when going from 4k to 16k tokens, and minimal gains for 2WikiMultihopQA (Figure~\ref{fig:all_scaling_data}). Full per-configuration results appear in Table~\ref{tab:scaling_metrics}.

\section{Discussion}\label{sec:discussion}

The results suggest a practical budget policy in our grid: allocate additional searches first (typically up to three), improve evidence quality via hybrid retrieval with re-ranking, and expand token generation limits primarily for multi-hop reasoning tasks. This ordering captures most of the observed gains while bounding compute.

One consequence of this framing is that the paper supports configuration decisions more directly than it supports claims about a universally best agent. Every result is conditioned on the same prompt scaffold, retriever stack, and budget rules. That makes the comparisons useful for deciding how to spend a fixed allowance, but it also means the paper should be read as a study of design leverage under control, not as a leaderboard over heterogeneous systems.

\subsection{Dataset-dependent Budget Leverage}
The most unusual effect of scaling the completion token budget appears on HotpotQA. While accuracy remains largely flat for TriviaQA and 2WikiMultihopQA, HotpotQA exhibits a sharp increase when the allowance rises from 4k to 16k tokens. We hypothesize that the datasets differ in the \emph{type} of difficulty they pose: 2WikiMultihopQA seems primarily constrained by retrieval, because locating the correct evidence is hard, whereas HotpotQA appears less retrieval-bound and more limited by the reasoning required to synthesise multiple pieces of evidence into a final response. This would explain why accuracy improves more distinctly with higher iterative search count on 2WikiMultihopQA compared to HotpotQA, whereas HotpotQA sees major gains for higher token count, unlike 2WikiMultihopQA.

These patterns also reveal an asymmetric relationship between search depth and context allocation. Accuracy increases monotonically with search steps across all models and datasets, with diminishing returns above three searches. However, context scaling shows that \textit{constrained completion tokens can outperform unlimited contexts when the model manages multiple searches}. This phenomenon is most evident on 2WikiMultihopQA, where GPT-4.1-mini achieves 55.84\% accuracy with 500 completion tokens but only 30.62\% with a single search at 16K tokens. Similarly, DeepSeek V3 reaches 27.68\% with 500 tokens versus 14.07\% with one search. Under tight token constraints, models adopt succinct behaviour, making efficient tool calls that enable multiple searches within budget. Conversely, unlimited single searches encourage verbose responses that consume the entire token budget without additional information gathering.

In our experiments, allocating budget toward search depth often outperforms allocating the same budget toward longer single-pass generations. For practitioners, moderately restrictive token budgets (500 to 2K) combined with multiple search opportunities can yield better accuracy than generous single-search configurations.

\subsection{The o4-mini Anomaly}
o4-mini presents the most puzzling behavioural pattern in our evaluation across different model capacities. Despite being among the top performers overall, it shows minimal response to most system enhancements. On HotpotQA, all enhancement strategies yield less than 1.10 percentage points improvement, a stark contrast to other models that see 4 to 18 point gains.

This resistance to enhancement may stem from o4-mini being a reasoning model with built-in chain-of-thought processing. As such, external planning components are often redundant. An exception appears on 2WikiMultihopQA, where combining planning with reflection produces a 25-point increase (38.12\% to 63.20\%). This suggests that while basic pre-planning may be redundant, reflection can create opportunities for mid-search strategy revision that complement o4-mini's internal reasoning process.

\subsection{Component Hierarchy and Model Variety}
The ablation results reveal that component effectiveness varies dramatically by model. Re-ranking with hybrid search provides the most consistent gains (6 to 18 percentage points for most models), making it the most reliable enhancement across model classes. However, the magnitude of improvement varies significantly by model, with smaller ones tending to benefit more from retrieval enhancements than larger ones.

Pre-planning strategies show a similar pattern: smaller models (LLaMA 3.1 8B, Qwen 3 14B) gain 4 to 12 percentage points from planning components, while larger models show more modest improvements. This suggests that external scaffolding becomes less valuable for models with stronger inherent capabilities. Combining pre-planning with reflection slightly degrades performance for some models, indicating that adaptive re-planning can be counter-productive; following an initial plan appears more effective than continuously revising strategy mid-search for conventional models.

\subsection{Deployment Recommendations}
Based on our evaluation trends, we recommend prioritising iterative search depth, followed by component selection, then token generation limits in budget allocation. For cost-sensitive deployments, LLaMA 3.1 8B (the smallest model evaluated) with three searches and BM25 already surpasses the single-search baselines of larger systems, except for o4-mini and GPT-4.1-mini. Adding pre-planning and reflection allows it to match or exceed every model limited to one search while remaining dramatically cheaper.

For maximum accuracy, o4-mini with both planning strategies reaches 63.20\% on 2WikiMultihopQA, the highest performance observed, but this requires targeted configuration for complex tasks. Models without native chain-of-thought capabilities benefit more from hybrid retrieval and re-ranking than from elaborate reasoning scaffolds, so enhancements should match the underlying capability profile.

The practical implication is that the first deployment question should usually be ``How many searches can we afford?'' rather than ``How large a completion window should we buy?'' In our grid, the clearest gains come from one additional well-targeted retrieval step or from better ranking of the retrieved evidence. Longer generations help chiefly when the retrieved material already contains the right pieces and the model still needs room to synthesize them.

\vspace{-6pt}
\section{Conclusion}\label{sec:conclusion}

We present \ac{BCAS}, a budget-aware evaluation harness for studying agentic retrieval under explicit search and token limits. Using six \acp{LLM} across TriviaQA, HotpotQA, and 2WikiMultihopQA, we quantify how search depth, retrieval configuration, and completion budgets trade off accuracy and cost.

Across our experiments, three patterns are consistent. First, increasing search depth improves accuracy up to a small cap in most settings, with diminishing returns after roughly three searches. Second, hybrid lexical+dense retrieval with lightweight re-ranking provides the largest average accuracy gains in our HotpotQA ablation grid. Third, larger completion budgets matter most on synthesis-heavy HotpotQA settings, while retrieval-easier settings show faster saturation.

These results provide a practical configuration order for budgeted deployments: expand search depth first, improve evidence quality through retrieval and re-ranking, and then raise completion budgets when synthesis demands justify the added cost. By making tool and token budgets explicit and logging usage outcomes, BCAS supports reproducible accuracy-cost analysis under realistic constraints.

\subsection{Future Directions}
Future work should add direct non-agentic baselines, broaden component ablations beyond HotpotQA, and expand evaluation to open-web and multimodal environments. A second direction is tighter runtime reporting, including queueing and throughput behavior under production concurrency constraints.

\section{Limitations}
Our study focuses on factual QA benchmarks over static benchmark corpora rather than open-web retrieval. The reported trends may not transfer directly to domains with rapid knowledge drift, multilingual corpora, multimodal evidence, or long-horizon interactive tasks.

The benchmarks also emphasize short final answers with known references. That is useful for controlled comparison, but it leaves out settings where success depends on citation fidelity, extended synthesis, interactive clarification, or user-specific constraints. A policy that is attractive for benchmark QA may therefore need different stopping rules or retrieval breadth in production workflows.

The full component ablation is reported on HotpotQA only. We do not claim that the exact ordering of component gains is universal across datasets; TriviaQA and 2WikiMultihopQA may shift the relative value of retrieval quality, search depth, and completion budget.

For each method, we use a single prompt template across model families instead of per-model prompt tuning. This improves control for comparative measurement but may under-optimize some models, so cross-model comparisons should be interpreted as sensitivity to a shared scaffold rather than each model's best-case performance.

Answer correctness is measured with a binary LLM judge at scale. Although our manual audit indicates high agreement, binary grading can miss partial correctness and reasoning quality nuances.

Cost and latency estimates depend on provider pricing, rate limits, and runtime settings. We therefore report token and search usage as stable budget proxies alongside cost estimates, and treat cross-provider cost comparisons as approximate.

We do not include a pure non-agentic single-pass retrieve-then-read baseline in this version. The presented comparisons therefore isolate differences among budgeted agentic policies and do not directly estimate the incremental value of multi-turn agency over one-shot retrieval.

We also do not report head-to-head results against RL-trained open-web search agents. Those systems differ in training regime and retrieval environment, so direct comparison would require aligned tools, corpora, and model training conditions.

\section{Data and Code Availability}
Code, evaluation harness configurations, prompts, and analysis scripts used in this study are available in the BCAS repository:
\url{https://github.com/kmccleary3301/BCAS_RAG}.
We also provide dataset split hashes and evaluation outputs needed to reproduce the reported plots and tables under the same budget settings.

\section{Bibliographical References}
\bibliographystyle{lrec2026-natbib}
\bibliography{bibliography}

@article{newcombe1998paired,
  author  = {Newcombe, Robert G.},
  title   = {Improved Confidence Intervals for the Difference Between Binomial Proportions Based on Paired Data},
  journal = {Statistics in Medicine},
  year    = {1998},
  volume  = {17},
  number  = {22},
  pages   = {2635--2650},
  doi     = {10.1002/(SICI)1097-0258(19981130)17:22<2635::AID-SIM954>3.0.CO;2-C}
}

@inproceedings{lewis2020rag,
  title={Retrieval-augmented generation for knowledge-intensive nlp tasks},
  author={Lewis, Patrick and Perez, Ethan and Piktus, Aleksandra and Petroni, Fabio and Karpukhin, Vladimir and Goyal, Naman and K{\"u}ttler, Heinrich and Lewis, Mike and Yih, Wen-tau and Rockt{\"a}schel, Tim and Riedel, Sebastian and Kiela, Douwe},
  booktitle={Advances in Neural Information Processing Systems},
  volume={33},
  pages={9459--9474},
  year={2020}
}

@inproceedings{izacard2021fid,
  author    = {Gautier Izacard and Edouard Grave},
  title     = {Leveraging Passage Retrieval with Generative Models for Open Domain Question Answering},
  booktitle = {Proceedings of the 16th Conference of the European Chapter of the Association for Computational Linguistics: Main Volume},
  pages     = {874--880},
  year      = {2021}
}

@inproceedings{yao2023react,
  title={ReAct: Synergizing Reasoning and Acting in Language Models},
  author={Yao, Shunyu and Zhao, Jeffrey and Yu, Dian and Du, Nan and Shafran, Izhak and Narasimhan, Karthik and Cao, Yuan},
  booktitle={International Conference on Learning Representations (ICLR)},
  year={2023}
}

@article{xia2025selfreasoning,
  title   = {Improving Retrieval Augmented Language Model with Self-Reasoning},
  author  = {Yuan Xia and Jingbo Zhou and Zhenhui Shi and Jun Chen and Haifeng Huang},
  journal = {Proceedings of the AAAI Conference on Artificial Intelligence},
  volume  = {39},
  number  = {24},
  pages   = {25534--25542},
  year    = {2025},
  doi     = {10.1609/aaai.v39i24.34743},
  url     = {https://ojs.aaai.org/index.php/AAAI/article/view/34743}
}

@article{wilson1927probable,
  title={Probable inference, the law of succession, and statistical inference},
  author={Wilson, Edwin B},
  journal={Journal of the American Statistical Association},
  volume={22},
  number={158},
  pages={209--212},
  year={1927},
  publisher={Taylor \& Francis}
}

@inproceedings{press2023selfask,
  title={Measuring and narrowing the compositionality gap in language models},
  author={Press, Ofir and Zhang, Muru and Min, Sewon and Schmidt, Ludwig and Smith, Noah A and Lewis, Mike},
  booktitle={Findings of the Association for Computational Linguistics: EMNLP 2023},
  year={2023}
}

@inproceedings{jiang2023flare,
  title={Active Retrieval Augmented Generation},
  author={Jiang, Zhengbao and Xu, Frank F and Gao, Luyu and Sun, Zhiqing and Liu, Qian and Dwivedi-Yu, Jane and Yang, Yiming and Callan, Jamie and Neubig, Graham},
  booktitle={Proceedings of the 2023 Conference on Empirical Methods in Natural Language Processing (EMNLP)},
  year={2023}
}

@inproceedings{asai2024selfrag,
  title={Self-RAG: Self-Reflective Retrieval-Augmented Generation},
  author={Asai, Akari and Wu, Zeqiu and Wang, Yizhong and Sil, Avirup and Hajishirzi, Hannaneh},
  booktitle={International Conference on Learning Representations (ICLR)},
  year={2024}
}

@article{yan2024crag,
  title={CRAG: Corrective Retrieval Augmented Generation},
  author={Yan, Shi-Qi and Gu, Jia-Chen and Zhu, Yun and Ling, Zhen-Hua},
  journal={arXiv preprint arXiv:2401.15884},
  year={2024}
}

@article{jin2025searchr1,
  title={Search-R1: Training LLMs to Reason and Leverage Search Engines with Reinforcement Learning},
  author={Jin, B. and Zeng, H. and Yue, Z. and Yoon, J. and Arik, S. and Wang, D. and Zamani, H. and Han, J.},
  journal={arXiv preprint arXiv:2503.09516},
  year={2025}
}

@article{jiang2025deepretrieval,
  title={DeepRetrieval: Hacking Real Search Engines and Retrievers with Large Language Models via Reinforcement Learning},
  author={Jiang, P. and Lin, J. and Cao, L. and Tian, R. and Kang, S. and Wang, Z. and Sun, J. and Han, J.},
  journal={arXiv preprint arXiv:2503.00223},
  year={2025}
}

@inproceedings{2wikimultihop,
    title = "Constructing A Multi-hop {QA} Dataset for Comprehensive Evaluation of Reasoning Steps",
    author = "Ho, Xanh  and
      Duong Nguyen, Anh-Khoa  and
      Sugawara, Saku  and
      Aizawa, Akiko",
    editor = "Scott, Donia  and
      Bel, Nuria  and
      Zong, Chengqing",
    booktitle = "Proceedings of the 28th International Conference on Computational Linguistics",
    month = dec,
    year = "2020",
    address = "Barcelona, Spain (Online)",
    publisher = "International Committee on Computational Linguistics",
    url = "https://aclanthology.org/2020.coling-main.580/",
    doi = "10.18653/v1/2020.coling-main.580",
    pages = "6609--6625",
}

@techreport{gemini25,
  title        = {{Gemini 2.5: Pushing the Frontier with Advanced Reasoning, Multimodality, Long Context, and Next Generation Agentic Capabilities}},
  author       = {{Gemini Team}},
  institution  = {Google DeepMind},
  year         = {2025},
  type         = {Technical Report},
  url          = {https://storage.googleapis.com/deepmind-media/gemini/gemini_v2_5_report.pdf},
  note         = {Accessed: 2025-07-01}
}

@article{singh2024agentic,
  title={Agentic Retrieval-Augmented Generation: A Survey on Agentic RAG},
  author={Singh, A. and others},
  journal={arXiv preprint arXiv:2501.09136},
  year={2024}
}

@misc{openai2023function,
  title={Function calling and other API updates},
  author={{OpenAI}},
  year={2023},
  howpublished={\url{https://platform.openai.com/docs/guides/function-calling}},
  note={Accessed: 2025-07-01}
}

@misc{anthropic2024research,
  title={Introducing Claude's Research Mode},
  author={{Anthropic}},
  year={2024},
  howpublished={\url{https://www.anthropic.com/research}},
  note={Accessed: 2025-07-01}
}

@misc{judgingllmasajudgemtbenchchatbot,
      title={Judging LLM-as-a-Judge with MT-Bench and Chatbot Arena}, 
      author={Lianmin Zheng and Wei-Lin Chiang and Ying Sheng and Siyuan Zhuang and Zhanghao Wu and Yonghao Zhuang and Zi Lin and Zhuohan Li and Dacheng Li and Eric P. Xing and Hao Zhang and Joseph E. Gonzalez and Ion Stoica},
      year={2023},
      eprint={2306.05685},
      archivePrefix={arXiv},
      primaryClass={cs.CL},
      url={https://arxiv.org/abs/2306.05685}, 
}

@article{Llama3_2024,
  title={The Llama 3 Herd of Models},
  author={Dubey, Abhimanyu and Jauhri, Abhinav and Pandey, Abhinav and Kadian, Abhishek and Al-Dahle, Ahmad and Letman, Aiesha and Mathur, Akhil and Schelten, Alan and Yang, Amy and Fan, Angela and others},
  journal={arXiv preprint arXiv:2407.21783},
  year={2024}
}

@InProceedings{TriviaQA_2017,
     author = {Joshi, Mandar and Choi, Eunsol and Weld, Daniel S. and Zettlemoyer, Luke},
     title = {TriviaQA: A Large Scale Distantly Supervised Challenge Dataset for Reading Comprehension},
     booktitle = {Proceedings of the 55th Annual Meeting of the Association for Computational Linguistics},
     month = {July},
     year = {2017},
     address = {Vancouver, Canada},
     publisher = {Association for Computational Linguistics},
}

@inproceedings{HotpotQA_2018,
  title={{HotpotQA}: A Dataset for Diverse, Explainable Multi-hop Question Answering},
  author={Yang, Zhilin and Qi, Peng and Zhang, Saizheng and Bengio, Yoshua and Cohen, William W. and Salakhutdinov, Ruslan and Manning, Christopher D.},
  booktitle={Conference on Empirical Methods in Natural Language Processing ({EMNLP})},
  year={2018}
}

@misc{paradedbgithub,
  title        = "ParadeDB",
  author       = "Philippe Noël ",
  howpublished = "\url{https://github.com/paradedb/paradedb}",
  year         = 2024,
  note         = "Accessed: 2024-7-01"
}

@article{Cover_Density,
    title = {Relevance ranking for one to three term queries},
    journal = {Information Processing \& Management},
    volume = {36},
    number = {2},
    pages = {291-311},
    year = {2000},
    issn = {0306-4573},
    doi = {https://doi.org/10.1016/S0306-4573(99)00017-5},
    url = {https://www.sciencedirect.com/science/article/pii/S0306457399000175},
    author = {Charles L.A. Clarke and Gordon V. Cormack and Elizabeth A. Tudhope}
}

@misc{bge-m3,
      title={BGE M3-Embedding: Multi-Lingual, Multi-Functionality, Multi-Granularity Text Embeddings Through Self-Knowledge Distillation}, 
      author={Jianlv Chen and Shitao Xiao and Peitian Zhang and Kun Luo and Defu Lian and Zheng Liu},
      year={2024},
      eprint={2402.03216},
      archivePrefix={arXiv},
      primaryClass={cs.CL},
      url={https://arxiv.org/abs/2402.03216}
}

@misc{baai_rerank,
      title={Making Large Language Models A Better Foundation For Dense Retrieval}, 
      author={Chaofan Li and Zheng Liu and Shitao Xiao and Yingxia Shao},
      year={2023},
      eprint={2312.15503},
      archivePrefix={arXiv},
      primaryClass={cs.CL},
      url={https://arxiv.org/abs/2312.15503}
}

@misc{gemma2,
      title={Gemma 2: Improving Open Language Models at a Practical Size}, 
      author={Gemma Team and Morgane Riviere and Shreya Pathak and Pier Giuseppe Sessa and Cassidy Hardin and Surya Bhupatiraju and Léonard Hussenot and Thomas Mesnard and Bobak Shahriari and Alexandre Ramé and others},
      year={2024},
      eprint={2408.00118},
      archivePrefix={arXiv},
      primaryClass={cs.CL},
      url={https://arxiv.org/abs/2408.00118}, 
}

@misc{GPT4_1_Announcement,
  author       = {OpenAI},
  title        = {Introducing GPT-4.1 in the API},
  year         = {2024},
  url          = {https://openai.com/index/gpt-4-1/},
  note         = {Accessed: 2025-07-04}
}

@misc{o3_System_Card,
  author       = {OpenAI},
  title        = {OpenAI o3 System Card},
  year         = {2025},
  url          = {https://openai.com/index/o3-o4-mini-system-card/},
  note         = {Accessed: 2025-06-23}
}

@misc{deepseek2024,
  author       = {DeepSeek},
  title        = {DeepSeek-V3 Technical Report},
  year         = {2024},
  url          = {https://github.com/deepseek-ai/DeepSeek-V3},
  note         = {Accessed: 2025-01-15}
}

@misc{qwen2024,
  author       = {Alibaba Cloud},
  title        = {Qwen Technical Report},
  year         = {2024},
  url          = {https://github.com/QwenLM/Qwen},
  note         = {Accessed: 2025-01-15}
}

\end{document}